\documentclass[11pt,a4paper]{article}
\usepackage{note}
\usepackage{bm}
\usepackage[margin=1in]{geometry}
\usepackage{color}
\usepackage{cite}
\usepackage{algorithm}
\usepackage{algpseudocode}
\usepackage{empheq}

\newtheorem{theorem}{\bf Theorem}
\newtheorem{lemma}{\bf Lemma}

\title{A Communication-Efficient Multi-Agent Actor-Critic Algorithm for Distributed Reinforcement Learning}

\author{Yixuan Lin, Kaiqing Zhang, Zhuoran Yang, Zhaoran Wang, Tamer Ba\c{s}ar, \\
Romeil Sandhu, and Ji Liu
\thanks{
Y. Lin is with the Department of Applied Mathematics and Statistics at Stony Brook University (\texttt{yixuan.lin.1@stonybrook.edu}).
K. Zhang and T. Ba\c{s}ar are with the Coordinated Science Laboratory  at University of Illinois at Urbana-Champaign (\texttt{\{kzhang66,basar1\}@illinois.edu}), and their research was supported in part by the US Army Research Laboratory (ARL) Cooperative Agreement W911NF-17-2-0196.
Z. Yang is with the Department of Operations Research and Financial Engineering at Princeton University
(\texttt{zy6@princeton.edu}).
Z. Wang is with the Department of Industrial Engineering and Management Sciences at Northwestern University
(\texttt{zhaoran.wang@northwestern.edu}).
R. Sandhu is with the Departments of Bioinformatics and Computer Science at Stony Brook University
(\texttt{romeil.sandhu@stonybrook.edu}).
J. Liu is with the Department of Electrical and Computer Engineering at Stony Brook University
(\texttt{ji.liu@stonybrook.edu}).
}
}

\begin{document}

\maketitle

\begin{abstract}

This paper considers a distributed reinforcement learning problem in which a network of multiple agents aim to cooperatively maximize the globally averaged return through communication with only local neighbors. A randomized communication-efficient multi-agent actor-critic algorithm is proposed for possibly unidirectional communication relationships depicted by a directed graph. It is shown that the algorithm can solve the problem for strongly connected graphs by allowing each agent to transmit only two scalar-valued variables at one time. 

\end{abstract}


\section{Introduction}

Recently, there has been increasing interest in developing distributed machine learning algorithms. Notable examples include distributed linear regression \cite{regression}, multi-arm bandit \cite{mab}, reinforcement learning (RL) \cite{kaiqing}, and deep learning \cite{deep}.
Such algorithms have promising applications in large-scale networks, such as social platforms, online economic networks, cyber-physical systems, and Internet of Things, primarily because in such a complex network, it is impossible to collect
all the information at the same point and each
component of the network may not be willing to share its private information due to privacy issues.


Multi-agent reinforcement learning (MARL) problems have recently received increasing attention.  In general, MARL  problems are investigated  in settings that are either collaborative, competitive, or a mixture of the two. For collaborative MARL, the most rudimentary  framework is the canonical multi-agent Markov decision process 
\cite{boutilier1996planning,lauer2000algorithm}, where the agents share a common reward function that is determined by the joint actions of all agents. Another notable framework for collaborative MARL is the team Markov game model, also with a shared reward function among agents \cite{littman2001value,wang2003reinforcement}.  These two frameworks  were then extended to the setting where agents are allowed to have heterogeneous reward functions \cite{kar2013cal,kaiqing,zhang2018finite,lee2018stochastic,doan2019convergence}, collaborating with the goal of maximizing the long-term return corresponding to the team averaged reward. In particular, these works focused on a \emph{fully-decentralized/distributed} setting, where there exists no central controller to coordinate the agents to achieve the overall team goal. Instead, the agents are connected via a communication network and are only able to exchange information with the neighbors on the network. There is  also an  ever-growing  number of  works on MARL in competitive and mixed settings \cite{hu2003nash,foerster2016learning,lowe2017multi,omidshafiei2017deep}, where most of the recent ones are empirical works without theoretical convergence guarantees. In this work, we focus on the decentralized/distributed and collaborative MARL setting with networked agents as in \cite{kar2013cal,kaiqing,zhang2018networked}.

Within this setting, the work of \cite{kaiqing} proposed the first fully distributed actor-critic algorithm.
The algorithm allows the agents to exchange information over a communication network with possibly sparse connectivity at each agent, which improves the scalability of the multi-agent model with a high population  of agents and thus tackles one of the long-standing challenges in general MARL problems \cite{shoham2003multi}. 
A detailed comparison of the problem setting to the other existing ones on multi-agent and collaborative MARL is provided in \cite{kaiqing}.


A possible communication issue of the algorithms proposed in \cite{kaiqing} is that they require each agent
to transmit the entire vector of its estimate of $\omega$ to its neighbors at each time.
Such communication-costly algorithms may not be possible to secure in some learning applications,
for example, when the size of $\omega$ is very large but each agent has limited communication capacity.
In this paper, we propose an approach in which each agent broadcasts
only one (scaled) entry of its estimate of $\omega$, thus significantly reducing communication costs at each iteration.
The algorithms in \cite{kaiqing} more or less rely on doubly stochastic matrices, which implicitly requires bidirectional communication between each pair of neighboring agents. This requirement restricts the applications of the algorithm in scenarios with possibly uni-directional communication. To get around this limitation, we propose a variant using the idea of push-sum \cite{pushsum} with which each agent only needs to transmit two scalar-valued variables at each time, and in particular, each agent can independently determine which entry of the estimate of $\omega$ to transmit. For simplicity, we take a randomized way and show that with this very cheap communication scheme, the algorithm can solve the distributed reinforcement learning problem for any strongly connected graph that depicts the communication relationships among the agents. 
Note that  in parallel with  this work, \cite{chen2018communication} appeared to be the first one that considers the communication efficiency in MARL and distributed RL in general. However, the setting in \cite{chen2018communication} assumes the existence of   a central controller. In contrast, to the best of our knowledge, our work proposes the first communication-efficient MARL/distributed RL algorithm for networked agents.

The remaining part of this paper is organized as follows.
Section II introduces the MARL problem and presents a communication-efficient variant of the algorithm in \cite{kaiqing}, which essentially requires bi-directional communication and transmission coordination between neighboring agents. Motivated by the restrictive requirements, Section III proposes a multi-agent actor-critic algorithm for uni-directional communication, based on which, a modification is proposed in Section IV, yielding a communication-efficient algorithm over directed graphs without any transmission coordination among the agents. The paper ends with some concluding remarks in Section V, followed by an appendix.

\section{Problem Formulation}\label{sec:back}

In this section, we introduce the background and formulation of the MARL problem with networked agents.

\subsection{Networked Multi-Agent MDP}\label{sec:marl}

Consider   a team of $N$ agents, denoted by $\cN = \{1,2,\ldots,N\}$, operating in a common environment. 
It is assumed that  no central controller that can either collect rewards or make the  decisions for the agents  exists. 
In contrast, the agents are connected  by a possibly time-varying  communication network depicted  by an undirected graph     $\cG_t=(\cN, \cE_t)$, where  the set of communication links   at time $t\in \NN$ is denoted by $\cE_t$. 
%
Then, a \emph{networked multi-agent MDP} model can be defined by  a  tuple 
$( \cS, \{\cA^i\}_{i\in\cN}, P, \{R^i\}_{i\in\cN},  \{ \cG_t \}_{t \geq 0} )$, where $\cS$ is the  state space shared by all the agents in $\cN$,  $\cA^{i}$ is the action space of  agent $i$, and $ \{ \cG_t \}_{t \geq 0}$ is a sequence of time-varying communication networks.  
For each agent $i$, 
$R^i :\cS\times\cA \to\RR$ is
the local reward function, where $\cA=\prod_{i=1}^{N}\cA^i$ is the joint action space.   $P:\cS\times\cA\times\cS\to[0,1]$ denotes  the state transition probability of the MDP.   It is assumed  throughout the paper that   the states 
are  globally observable and but  the  rewards are observed only  locally.

The networked multi-agent MDP evolves as follows.  Each agent $i$ chooses its own action $a^i_t$  given state $s_t$ 
at time $t$, according to a local policy, i.e., the probability of choosing action $a^i$ at state $s$, $\pi^i :\cS\times\cA^i\to[0,1]$. 
Note that the joint policy of all agents,  $\pi \colon \cS\times\cA\to[0,1]$, satisfies   $\pi(s,a)=\prod_{i\in \cN} \pi ^i(s,a^i) $.  
Also, a reward $r^i_{t+1}$ is received by agent $i$ after executing the action. 
%
To make the search of the optimal joint policy tractable,   we assume that the  local policy is parameterized  by $\pi_{\theta^i}^i$, where   $\theta^i\in   {\Theta} ^ i$ is the parameter, and      $\Theta^i\subseteq \RR^{m_i}$ is a compact set.  The parameters are concatenated as  $\theta=[(\theta^1)^{\top}, \cdots, (\theta^N)^{\top}]^{\top}\in\Theta$, where $\Theta=\prod_{i=1}^N\Theta^i$. The joint policy is thus given by 
$ \pi_{\theta} (s, a) = \prod_{i\in \cN} \pi_{\theta^i} ^i (s, a_i).$ 
We first make a standard regularity assumption on the model and the policy  parameterization.

\begin{assumption}
\label{assum:MDP_Erg}
For any $i \in \cN$,  $s \in \cS$,  and $a^i\in\mathcal{A}^i$, the policy function $\pi_{\theta^{i} }^i(s,a^i)>0$ for any $\theta^{i}\in\Theta^i$. Also,   $\pi_{\theta^{i} }^i(s,a^i)$ is continuously differentiable with respect to the parameter $\theta^i$ over $\Theta^i$.   
In addition, 
for any $\theta \in \Theta$, let $P^{ {\theta}}$ be  the transition matrix of the Markov chain $\{s_t\}_{t\geq 0}$  induced by policy $\pi_{\theta}$, that is, for any $s, s' \in \cS$
\#\label{equ:Transi_Def}
P^{\theta}(s' \given s) = \sum_{a \in \cA} \pi_{\theta} (s, a) \cdot P( s' \given s, a).
\#

\noindent We assume that the Markov chain  $\{s_t\}_{t\geq 0}$
is irreducible and aperiodic under any $\pi_{\theta}$, with the stationary distribution denoted by $d_{\theta}$.
\end{assumption}

Assumption \ref{assum:MDP_Erg} has been made in the existing work on actor-critic (AC)  algorithms with function approximation   \cite{konda2000actor,bhatnagar2009natural}.
It  implies that the Markov chain of the  state-action pair $\{(s_t,a_t)\}_{t\geq 0}$  has a  stationary distribution $d_{\theta}(s)\cdot\pi_{\theta}(s,a)$ for any  $s\in\cS$ and $a\in\cA$.

The  objective of the agents is to collaboratively find a  policy $\pi_{\theta}$ that maximizes the \emph{globally} averaged long-term  return  over the network based  solely on  \emph{local} information, namely,  
\#\label{eq:rewards}
\max_{\theta}~~J({\theta})=&~\lim_{T}~~\frac{1}{T}\EE\bigg(\sum_{t=0}^{T-1} \frac{1}{N}\sum_{i\in\cN} r^i_{t+1}\bigg)\nonumber\\
= &~\sum_{s\in\cS,a\in\cA}d_{\theta}(s)\pi_{\theta}(s,a) \cdot \overline R(s,a), 
\#

\noindent where $\overline R(s, a) = N^{-1} \cdot \sum_{i\in\cN} R^i (s,a)$ is the globally  averaged reward function.
Let $\overline{r}_t={N}^{-1} \cdot \sum_{i\in\cN} r^i_t $; then,  we have  $\overline R(s,a)=\EE[\overline{r}_{t+1}\given s_t=s,a_t=a]$.
Thus, the global relative action-value  function   
under policy $\pi_{\theta}$ can be defined accordingly as
$$\label{equ:MA_Q_def}
Q_{{\theta}}(s,a)=\sum_{t}\EE \bigl [\overline r_{t+1}-J({\theta})\given s_0=s,a_0=a,\pi_{\theta} \bigr], 
$$

\noindent and the global relative  state-value function $V_{{\theta}}(s)$ is defined as $V_{{\theta}}(s)=\sum_{a\in\cA}\pi_\theta(s,a)Q_{{\theta}}(s,a)$.
For simplicity, hereafter we will refer to $V_{\theta}$ and $Q_{\theta}$  as \emph{state-value} function  and \emph{action-value} function only. Furthermore, the \emph{advantage function} can be defined as $A_{\theta}(s,a)=Q_{\theta}(s,a)-V_{\theta}(s)$.


As the basis for developing multi-agent actor-critic algorithms, the following policy gradient theorem was established in \cite{kaiqing} for MARL.



{\em Policy Gradient Theorem for MARL:}
[Theorem $3.1$ in \cite{kaiqing}] \label{thm:policy_grad}
		For any $\theta \in \Theta$ and  
		 any agent $i \in \cN$, we  define the local advantage function $A_{\theta}^i \colon \cS \times \cA \rightarrow \RR$ as 
		\#\label{eq:local_advantage}
		A^i_{{\theta}}(s,a)=&~Q_{{\theta} } (s, a)-\tilde{V}_{ {\theta}}^i(s,a^{-i}), 
		\#
\noindent {where} $\tilde{V}_{ {\theta}}^i(s,a^{-i})=\sum_{a^i\in\cA^i}\pi _{\theta^i}^i (s,a^i) \cdot Q_{{\theta} } (s, a^i,a^{-i})$. 
		We use $a^{-i}$ to denote the  actions of all agents except for $i$.
Then, the gradient of $J(\theta)$ with respect to $\theta^i$ is given by
		\#\label{eq:policy_gradient_thm}
		\nabla _{\theta^{i} } J({\theta})
		&=\EE _{s \sim d_{{\theta}}, a \sim \pi_{\theta} } \left [ \nabla _{\theta^i} \log \pi _{\theta^i}^i (s,a^i)\cdot  A_{{\theta}}(s,a) \right ]\nonumber\\
		&=\EE _{s \sim d_{{\theta}}, a \sim \pi_{\theta} } \left [ \nabla _{\theta^i} \log \pi _{\theta^i}^i (s,a^i)\cdot  A^i_{{\theta}}(s,a) \right ].
		\#


\subsection{A Motivating Algorithm} \label{sec:sub_sec_algo}

As mentioned earlier, the distributed actor-critic algorithms proposed in \cite{kaiqing} require each agent to transmit its estimate of $\omega$ to all its neighbors at each time, which can be communication-expensive when the size of $\omega$ is very large. A natural idea to reduce the communication cost at each time step is to allow each agent to transmit only partial entries of its estimate vector to its neighbors at one time. Such an idea has been explored in distributed algorithms for solving linear algebraic equations \cite{xiaobin} and finding a common fixed point among a family of nonlinear maps \cite{daniel}. 

We first propose a communication-efficient algorithm based on the algorithm in \cite{kaiqing} and then show its limitation in implementation, which serves as a motivating algorithm for our main contribution in the next sections. 

The algorithm is 
based on the local advantage function $ A_{\theta}^i$ defined in \eqref{eq:local_advantage}, which requires estimating  the  action-value function $Q_ {\theta}  $ of  policy $\pi_{\theta}$.
Consider $Q(\cdot,\cdot; \omega ) \colon \cS \times \cA \rightarrow \RR$, a family of functions parametrized by $\omega \in \RR^K$, where $K\ll|\cS| \cdot |\cA|$. 
It is assumed that each agent $i$ maintains its own parameter 
$\omega^i $  and uses $Q(\cdot,\cdot; \omega ^i) $
 as a local estimate of  $Q_{\theta}$. 

The algorithm consists of two steps,  the \emph{actor} step and the \emph{critic} step.   In the critic step, an update based on \emph{temporal difference (TD) learning} is performed at each agent to estimate  $Q(\cdot,\cdot; \omega ^i)$, followed by a linear combination of its neighbors' parameter estimates. 
Different from Algorithm 1 in \cite{kaiqing} in which the parameter sharing step is the consensus update for all $N$ agents' entire vectors of their estimates of $\omega$, the algorithm here allows each agent to randomly transmit some entries of its estimate vector.
For simplicity, suppose that each agent randomly picks one entry of its estimate vector at each time and then transmits the entry to its neighbors. 
Then, the critic step involves a consensus update for each entry $k$ of the estimate vector, depending on which agents transmit the corresponding entry at time $t$, which involves a weight matrix $C^k_t=[c^k_t(i,j)]_{N\times N}$, where $c^k_t(i,j)$ is the  weight on the $k$-th entry transmitted from agent $j$ to agent $i$ at time $t$. 
Specifically, the critic step  iterates as follows: 
\begin{empheq}[left = \empheqlbrace]{align}
\mu^i_{t+1}&=(1-\beta_{\omega,t})\cdot\mu^i_t+\beta_{\omega,t} \cdot r^i_{t+1},\notag\\ 
\tilde{\omega}^i_{t}&=\omega^i_{t}+\beta_{\omega,t}\cdot \delta^i_{t}\cdot \nabla_{\omega} Q_t(\omega^i_t),\label{equ:MARL_critic_1}\\ 
{\omega}^{ik}_{t+1}&=\sum_{j\in\cN}c^k_t(i,j)\cdot \tilde{\omega}^{jk}_{t},\;\; k\in\{1,2,\ldots,K\},\notag
\end{empheq}
where ${\omega}^{ik}_{t}$ and $\tilde{\omega}^{ik}_{t}$ denote the $k$-th entry of ${\omega}^{i}_{t}$ and $\tilde{\omega}^{i}_{t}$, respectively,
$\mu^i_t$ tracks the long-term return of agent $i$, $\beta_{\omega, t} > 0$ is the stepsize,    $Q_t(\omega)=Q(s_t,a_t; \omega)$ for any $\omega$, and the local  \emph{action-value TD-error} $\delta^i_{t}$ in \eqref{equ:MARL_critic_1} is defined as
\#\label{equ:Local_TD_def}
\delta^i_{t}=r^i_{t+1}-\mu^i_{t}+ Q_{t+1}(\omega^i_t)-Q_t(\omega^i_t).
\#
The actor step is motivated by \eqref{eq:policy_gradient_thm} and is the same as that of Algorithm~1 in \cite{kaiqing}.

In the special case when all weight matrices $C^k_t=[c^k_t(i,j)]_{N\times N}$ are the same, i.e., $C^k_t= C_t=[c_t(i,j)]_{N\times N}$, $k\in\{1,\ldots,K\}$,
the algorithm simplifies to Algorithm~1 in \cite{kaiqing}, i.e., all agents' estimate vectors are transmitted simultaneously at each time. 
For this case, it has been shown in \cite{kaiqing} that the algorithm will guarantee the convergence upon the following assumption on the weight matrix $C_t$.

\begin{assumption}\label{assum:MARL_weighted_matrix}
The sequence of random matrices $\{ C_t \}_{t\ge0}$ satisfies the following conditions:
\begin{enumerate}
\item $C_t$ is row stochastic and $\mathbb{E}(C_t)$ is column stochastic.\footnote{
A nonnegative matrix is called row (or column) stochastic if all its row (or column) sums equal to one.}
There exists a constant $\eta > 0$ such that $c_t(i,j) \ge \eta$ for any $c_t(i,j) > 0$.
\item $C_t$ respects the communication graph $\cG_t$, i.e., $c_t(i,j) = 0$, if $(j,i) \not \in \mathcal E_t$.
\item The spectral norm of $\mathbb{E}[C_t^\top \cdotp (I-\1\1^\top/N) \cdotp C_t ]$ is strictly less than one.
\item Given the $\sigma$-algebra generated by the random variable before time $t$, $C_t$ is conditionally independent of $r_{t+1}^i$ for any $i \in \mathcal{N}$.
\end{enumerate}
\end{assumption}
Here $\1_N$ denotes the $N$-dimensional column vector whose entries are all equal to one.

We now consider the heterogeneous $C^k_t$ case proposed in our algorithm. 
Let $\omega = [(\omega^{1})^\top,\dots, (\omega^{N})^\top]^\top$ and $\tilde\omega = [(\tilde\omega^{1})^\top,\dots, (\tilde\omega^{N})^\top]^\top$.
Then, from~\rep{equ:MARL_critic_1}, it can be verified that 
$$
\omega_{t+1} = \bar C_{t} \cdotp \tilde \omega_{t},
$$
where $\bar C_t  =\sum_{k=1}^K C^k_{t} \otimes (e_k e_k^\top) \in \mathbb{R}^{NK\times NK}$ and $e_k $ is $k$-th unit vector. 
More can be said.

\begin{proposition}\label{yixuan}
\label{lemma:MARLone_weighted_matrix_assum}
If $C_t^k$ satisfies Assumption \ref{assum:MARL_weighted_matrix} for all $k\in\{1,\ldots,K\}$, then $\bar C_t$ does as well.
\end{proposition}

The proof of this proposition can be found in the appendix.

The above proposition implies that the proposed communication-efficient algorithm can also guarantee the convergence if each entry's weight matrix $C^k_t$ satisfies Assumption~\ref{assum:MARL_weighted_matrix}.\footnote{
The proof is essentially the same as that in \cite{kaiqing}.} 
However, satisfying the assumption leads to restriction in implementation. Specifically, there only exist three distributed approaches to implement a weight matrix satisfying the assumption -- the Metropolis algorithm \cite{xiao2005scheme}, pairwise gossiping \cite{boyd2006randomized}, and broadcast gossip \cite{aysal2009broadcast}. However, the broadcast gossip cannot always guarantee average consensus and thus cannot be applied to the distributed RL problem considered here, and both the Metropolis algorithm and pairwise gossiping require bi-directional communication between each pair of neighboring agents. 
Thus, the implementation of the above communication-efficient algorithm requires bi-directional communication, i.e., the communication graph $\mathcal G_t$ must be undirected. Moreover, when an agent $i$ receives the $k$-th entry from its neighbor $j$'s estimate, agent $i$ must send its $k$-th entry to agent $j$ as well. Since each agent randomly picks one of its entries, in the worst scenario when each agent picks a distinct entry, the above requirement may lead all the agents to transmit $N$ entries at one time. 
To get around this limitation, we will propose a multi-agent actor-critic algorithm for directed graphs in the next section, and then modify it to be communication-efficient in Section IV.

\section{Multi-Agent Actor-Critic over Directed Graphs}

In this section, we propose a multi-agent actor-critic algorithm for directed communication graphs by exploiting the idea of the push-sum protocol \cite{kempe}. 
For simplicity, we focus on fixed graphs which are assumed to be strongly connected. The algorithms and their convergence results can be extended to time-varying directed graphs with a certain mild condition on joint strong connectivity.

Let each agent $i$ have control over a scalar-valued variable $y^{i}_t$ whose initial value $y^{i}_0=1$.
The critic step of the algorithm iterates as follows:
\begin{empheq}[left = \empheqlbrace]{align}
\mu^i_{t+1}&=(1-\beta_{\omega,t})\cdot\mu^i_t+\beta_{\omega,t} \cdot r^i_{t+1},\notag\\ 
\tilde{\omega}^i_{t}&=\omega^i_{t}+\beta_{\omega,t}\cdot \tilde \delta^i_{t}\cdot \nabla_{z} Q_t(z^i_t),\notag\\ 
{\omega}^i_{t+1}&=\sum_{j\in\cN}b(i,j)\cdot \tilde{\omega}^j_{t},\notag\\
y^i_{t+1}&=\sum_{j\in\cN}b(i,j)\cdot y^j_{t},\label{equ:push_critic_1}\\
z^i_{t+1}&=\frac{{\omega}^i_{t+1}}{y^i_{t+1}},\notag
\end{empheq}
\noindent where $\mu^i_t$ tracks the long-term average return of agent $i$, $\beta_{\omega, t} > 0$ is the stepsize,  $Q_t(z)$ denotes $Q(s_t,a_t; z)$ for any $z$, 
and $b(i,j)$ will be specified shortly. 
The local  \emph{action-value TD-error} $\tilde \delta^i_{t}$ in \eqref{equ:push_critic_1} is given by
\#\label{equ:push_TD_def}
\tilde \delta^i_{t}=r^i_{t+1}-\mu^i_{t}+ Q_{t+1}(z^i_t)-Q_t(z^i_t).
\#
As for the actor step, each agent $i$  improves its policy via
\#\label{equ:push_actor_1}
	\theta^i_{t+1}=\theta^i_{t}+\beta_{\theta,t}\cdot A^i_t\cdot\psi^i_t,
\#
where $\beta_{\theta, t} >0$ is the stepsize, and $A^i_t$ and $\psi^i_t $ are defined as
\#
A^i_t &=  Q_t(z^i_t)-\sum_{a^i\in\cA^i} \pi_{\theta^i_t}^i(s_{t},a^i)\cdot Q(s_t,a^i,a^{-i}_t; z^i_t),\notag\\  \psi^i_t &= \nabla_{\theta^i}\log\pi_{\theta^i_t}^i(s_t,a^i_t).\label{equ:push_Advan_def}
\#
Let $\cG=(\cN, \cE)$ be the underlying communication graph. Then, $b(i,j)$ is given as follows: For all $i,j\in\cN$,
\begin{align*}
b(i,j)=&~ (1+d_j)^{-1} , ~~\text{if } (j,i)\in\mathcal{E}, \notag\\
b(i,j)=&~0, ~~\;\;\;\;\;\;\;\;\;\;\;\;\;\;\;\text{if } (j,i)\not\in\mathcal{E},
\end{align*}
where 
$d_j$ is the number of out-going neighbors\footnote{
Suppose that $(i,j)$ is a directed edge in graph $\cG$. We say that agent $i$ is an in-neighbor of agent $j$, and that agent $j$ is an out-neighbor of agent $i$.} 
of agent $j$, or equivalently, the out-degree of vertex $j$ in $\mathcal G$.
Let $B$ be the matrix whose $ij$-th entry is $b(i,j)$. Then, it is easy to see that 
\begin{enumerate}
    \item $B$ is column stochastic, and there exists a constant $\eta>0$ such that $b(i,j) \ge \eta$ for any $b(i,j) > 0$;
    \item $B$ respects the communication graph $\cG$; 
    \item Given the $\sigma$-algebra generated by the random variable before time $t$, $B$ is conditionally independent of $r_{t+1}^i$ for any $i \in \mathcal{N}$.
\end{enumerate}

We impose the following assumptions for the actor-critic algorithm which are either mild or standard; see \cite{kaiqing} for detailed discussions on these assumptions.


\begin{assumption}\label{assum:Data_Bounded}
The instantaneous  reward  $r^i_t$ is uniformly bounded for any $i\in\cN$ and $t\geq 0$.
\end{assumption}


\begin{assumption}
\label{assum:Stepsizes}
The stepsizes $\beta_{\omega,t}$ and $\beta_{\theta,t}$ satisfy
\$
&\sum_{t}\beta_{\omega,t}=\sum_{t}\beta_{\theta,t}=\infty,\\
&\sum_{t}\beta^2_{\omega,t}+\beta^2_{\theta,t}<\infty.
\$
In addition, $\beta_{\theta,t}=o(\beta_{\omega,t})$, and $\lim_{t}{\beta_{\omega,t+1}}\cdot{\beta^{-1}_{\omega,t}}=1$.
\end{assumption}


\begin{assumption}
\label{assum:Linear_Feature_Alg_1}
For each agent $i$, the  function $Q(s,a;z)$ is parametrized as $Q(s,a;z)=z^{\top}\phi(s,a)$, where $\phi(s,a)=[\phi_1(s,a),\cdots,\phi_K(s,a)]^\top\in\mathbb{R}^K$ is the feature associated with  $(s,a)$. The feature vector $\phi(s,a)$ is  uniformly bounded for any $s\in\cS, a\in\cA$.
Furthermore,  
the feature matrix $\Phi\in\RR^{|\cS|\cdot|\cA|\times K}$ has  full column rank, where the $k$-th column  of $\Phi$ is $[\phi_k(s,a), s\in\cS, a\in\cA]^\top$ for any $k \in [K]$.   Also, for any $u\in\RR^K$, $\Phi u\neq \1_K$.
\end{assumption}

\begin{assumption}\label{assum:Policy_Proj}
The update of the policy parameter $\theta^i_{t}$ includes a local projection operator, $\Gamma^i: \RR^{m_i}\to\Theta^i\subset \RR^{m_i}$, that projects any $\theta^i_{t}$ onto the compact set $\Theta^i$. Also, we assume that   $\Theta=\prod_{i=1}^{N}\Theta^i$  is large enough to include at least one local minimum of $J({\theta})$.
\end{assumption}



For simplicity,   we define  $P^{{\theta}}(s',a'\given s,a)=P(s'\given s,a)\pi_{\theta}(s',a')$\footnote{With slight abuse of notation, the expression $P^{{\theta}}$ has the same form as the transition probability matrix of the Markov chain $\{s_t\}_{t\geq 0}$ under policy $\pi_{\theta}$ (see the definition in \eqref{equ:Transi_Def}).  These two matrices can be easily differentiated by the context.},  
$\Db^{s,a}_{\theta}=\diag[d_{\theta}(s) \cdot \pi_{\theta} (s,a), s\in\cS, a\in\cA]$, and  $\overline{R}=[\overline{R}(s,a),s\in\cS,a\in\cA]^\top\in\RR^{|\cS|\cdot|\cA|}$.
We then define the operator $T^Q_{\theta}:\RR^{|\cS|\cdot|\cA|}\to \RR^{|\cS|\cdot|\cA|}$ for any action-value vector $Q\in\RR^{|\cS|\cdot|\cA|}$ as
\#\label{equ:def_T_operat}
T^Q_{\theta}(Q)=\overline{R}-J({\theta})\cdot \1_{|\cS|\cdot|\cA|}+P^{{\theta}}Q.
\# 
We also define the vector $\hat \Gamma^i(\cdot)$ as 
\#\label{equ:hat_Gamma}
\hat \Gamma^i[g(\theta)] = \lim_{0<\eta\rightarrow0}\{ \Gamma^i[\theta^i+\eta \cdotp g(\theta)] - \theta^i\}/\eta
\#
for any $\theta \in \Theta$ and $g:\Theta \rightarrow \mathbb{R}^{\sum_{i\in \mathcal{N}}m_i}$ a continuous function. In case the limit above is not unique, 
$\hat \Gamma^i[g(\theta)]$ is defined as the set of all possible limit points of \eqref{equ:hat_Gamma}.

With the above notation, we have the convergence  of the critic step \eqref{equ:push_critic_1} -- \eqref{equ:push_TD_def} and actor step \rep{equ:push_actor_1} -- \rep{equ:push_Advan_def} 
given policy $\pi_{\theta}$ as follows.

\begin{theorem}
\label{thm:critic_conv_push}
Suppose that Assumptions \ref{assum:MDP_Erg} and    \ref{assum:Data_Bounded}-\ref{assum:Linear_Feature_Alg_1} hold, and that communication graph $\mathcal G$ is strongly connected. 
Then, for any given policy $\pi_{\theta}$, with the sequences $\{\mu^i_t\}$ and $\{z^i_t\}$ generated  from  \eqref{equ:push_critic_1} and \eqref{equ:push_TD_def}, we have  $\lim_t \sum_{i\in\cN}\mu^i_t\cdot N^{-1}= J(\theta)$ and $\lim_t z^i_t=\omega_\theta$ almost surely for any $i\in\cN$, where $
J(\theta)
$ is the globally  averaged return as defined in \eqref{eq:rewards}, and  $\omega_\theta$ is the unique solution to
$$
\label{equ:TD_Consens_Solu_push}
\Phi^{\top}\Db^{s,a}_{{\theta}}\big[T^Q_{\theta}(\Phi \omega_\theta)-\Phi \omega_\theta\big]={0}.
$$
Suppose further that Assumption \ref{assum:Policy_Proj} holds. 
Then, the sequence $\{\theta^i_t\}$ obtained from  \eqref{equ:push_actor_1} converges almost surely  to a  point in the set 
of the asymptotically stable equilibria of 
$$
\label{equ:theta_ODE_push}
\dot{\theta}^i=\hat{\Gamma}^i\big[\EE_{s_t \sim d_{{\theta}}, a_t \sim \pi_{\theta} }\big(A^i_{t,\theta}\cdot\psi^i_{t,\theta}\big)\big], ~~{\forall~~} i\in\cN.
$$
\end{theorem}

\vspace{.1in}

The algorithm in this section works for directed communication graphs, but still requires each agent to transmit its entire estimate vector. 
In the next section, we will modify the algorithm to significantly reduce the communication cost at each time. The theorem just stated is a special case of the theorem in the next section.

\section{A Communication-Efficient Algorithm over Directed Graphs}

In this section, we present a communication-efficient distributed actor-critic algorithm, modified from the algorithm in the last section, in which each agent can independently transmit one scaled entry of its state vector, and in total only transmit two scalars at each iteration.

In our communication-efficient algorithm, each agent $i$ randomly picks one of its entries, $k$, of its estimate vector at time $t$ with probability $p^{ik}_t$, and then transmit its scaled value to its out-neighbors. 
For simplicity, we assume that $p^{ik}_t=p^{ik}$ for all $t$, and that $p^{ik}>0$ and $\sum_{k=1}^K p^{ik} = 1$.
For each entry $k$, each agent $i$ has control over a scalar-valued variable $y^{ik}_t$ whose initial value is $y^{ik}_0=1$.

The critic step iterates as follows:
\begin{empheq}[left = \empheqlbrace]{align}
\mu^i_{t+1}&=(1-\beta_{\omega,t})\cdot\mu^i_t+\beta_{\omega,t} \cdot r^i_{t+1},\notag\\ 
\tilde{\omega}^i_{t}&=\omega^i_{t}+\beta_{\omega,t}\cdot \tilde \delta^i_{t}\cdot \nabla_{z} Q_t(z^i_t),\notag\\ 
{\omega}^{ik}_{t+1}&=\sum_{j\in\cN}b^k_t(i,j)\cdot \tilde{\omega}^{jk}_{t},\;\; k\in\{1,2,\ldots,K\},\notag\\
y^{ik}_{t+1}&=\sum_{j\in\cN}b^k_t(i,j)\cdot y^{jk}_{t},\;\; k\in\{1,2,\ldots,K\},
\label{equ:pushone_critic_1}\\
z^{ik}_{t+1}&=\frac{{\omega}^{ik}_{t+1}}{y^{ik}_{t+1}},\notag
\end{empheq}
where ${\omega}^{ik}_{t}$ and $\tilde{\omega}^{ik}_{t}$ denote the $k$-th entry of ${\omega}^{i}_{t}$ and $\tilde{\omega}^{i}_{t}$, respectively, $\mu^i_t$ tracks the long-term return of agent $i$, $\beta_{\omega, t} > 0$ is the stepsize,   $Q_t(z)=Q(s_t,a_t; z)$ for any $z$,
and $b^k_t(i,j)$ will be specified shortly. 
The local  \emph{action-value TD-error} $\tilde \delta^i_{t}$ in \eqref{equ:pushone_critic_1} is given by
\#\label{equ:pushone_TD_def}
\tilde \delta^i_{t}=r^i_{t+1}-\mu^i_{t}+ Q_{t+1}(z^i_t)-Q_t(z^i_t).
\#
As for the actor step, each agent $i$  improves its policy via
\#\label{equ:pushone_actor_1}
	\theta^i_{t+1}=\theta^i_{t}+\beta_{\theta,t}\cdot A^i_t\cdot\psi^i_t,
\#
where $\beta_{\theta, t} >0$ is the stepsize. Moreover, $A^i_t$ and $\psi^i_t $ are defined as
\begin{align*}
A^i_t &=  Q_t(z^i_t)-\sum_{a^i\in\cA^i} \pi_{\theta^i_t}^i(s_{t},a^i)\cdot Q(s_t,a^i,a^{-i}_t; z^i_t),\notag\\  \psi^i_t &= \nabla_{\theta^i}\log\pi_{\theta^i_t}^i(s_t,a^i_t).\label{equ:pushone_Advan_def}
\end{align*}
Let $\cG=(\cN, \cE)$ be the underlying communication graph. Then, $b^k_t(i,j)$ is given as follows. For all $i,j\in\cN$,
$b^k_t(i,j)=(1+d_j)^{-1}$ if $(j,i)\in\mathcal{E}$ and agent $i$ picks $k$-th entry at time $t$; otherwise, $b^k_t(i,j)=0$,
where 
$d_j$ is the number of out-going neighbors of agent $j$, 
or equivalently, the out-degree of vertex $j$ in $\mathcal G$.
Let $B^k_t$ be the matrix whose $ij$-th entry is $b^k_t(i,j)$. Then, it is easy to see that
\begin{enumerate}
    \item Each $B^k_t$ is column stochastic, and there exists a constant $\eta>0$ such that $b^k_t(i,j) \ge \eta$ for any $b^k_t(i,j) > 0$;
    \item Each $B^k_t$ respects the communication relationship, i.e., $b^k_t(i,j) = 0$ if agent $i$ does not receive information from agent $j$ at time $t$;
    \item Given the $\sigma$-algebra generated by the random variable before time $t$, each $B^k_t$ is conditionally independent of $r_{t+1}^i$ for any $i \in \mathcal{N}$.
\end{enumerate}


It is worth emphasizing that in our communication-efficient algorithm, each agent $j$ only needs to transmit two scalars,
$\frac{\tilde{\omega}^{jk}_{t}}{1+d_j}$ and $\frac{y^{jk}_{t}}{1+d_j}$,
at each iteration, and all the computations are in a fully distributed manner.

\begin{theorem}\label{thm:critic_conv_pushone}
Suppose that Assumptions \ref{assum:MDP_Erg} and    \ref{assum:Data_Bounded}-\ref{assum:Linear_Feature_Alg_1} hold, and that communication graph $\cG$ is strongly connected. 
Then, for any given policy $\pi_{\theta}$, with the sequences $\{z^i_t\}$ and $\{\mu^i_t\}$ generated  from  \eqref{equ:pushone_critic_1} and \eqref{equ:pushone_TD_def}, we have  $\lim_t \sum_{i\in\cN}\mu^i_t\cdot N^{-1}= J(\theta)$ and $\lim_t z^i_t = \omega_\theta$ almost surely for any $i\in\cN$, where $
J(\theta)
$ is the globally  averaged return as defined in \eqref{eq:rewards}, and  $\omega_\theta$ is the unique solution to
$$\label{equ:TD_Consens_Solu_pushone}
\Phi^{\top}\Db^{s,a}_{{\theta}}\big[T^Q_{\theta}(\Phi \omega_\theta)-\Phi \omega_\theta\big]={0}.
$$
Suppose further that Assumption \ref{assum:Policy_Proj} holds. 
Then, the sequence $\{\theta^i_t\}$ obtained from  \eqref{equ:pushone_actor_1} converges almost surely  to a  point in the   set 
of the asymptotically stable equilibria of 
$$
\dot{\theta}^i=\hat{\Gamma}^i\big[\EE_{s_t \sim d_{{\theta}}, a_t \sim \pi_{\theta} }\big(A^i_{t,\theta}\cdot\psi^i_{t,\theta}\big)\big], ~~{\forall~} i\in\cN.
$$
\end{theorem}

\vspace{.1in}

To prove the above theorem, we need the following concepts and results. 

Define the operator $\langle \cdot \rangle : \mathbb{R}^{KN} \rightarrow \mathbb{R}^{K}$ as
$$ \langle x \rangle = \frac{1}{N} (\1_N^\top \otimes I_K) x= \frac{1}{N}\sum_{i\in \mathcal{N}}x^i$$
for any $x = [(x^{1})^\top,\dots, (x^{N})^\top]^\top\in\mathbb{R}^{KN}$ with $x^i\in\mathbb{R}^K$ for all $i\in\cN$.

\begin{lemma}\label{lemma:pushone_converge}
For all $i\in\cN$, $\lim_{t \rightarrow \infty}z^i_t = \lim_{t \rightarrow \infty}\langle \omega_t \rangle =\lim_{t \rightarrow \infty}\langle z_t \rangle$. 
\end{lemma}

\vspace{.08in}

{\em Proof:}
From the update of each agent $i$ in \rep{equ:pushone_critic_1},
\small
\begin{align*}
z_{t+1}^{ik} 
& = \frac{\omega_{t+1}^{ik}}{y_{t+1}^{ik}} \\
& = \frac{\sum_{j \in \mathcal{N}} b^k_t(i,j) \cdotp \omega_t^{jk} + b^k_t(i,j) \cdotp \beta_{\omega,t} \cdotp \tilde u_{t+1}^{jk}} {\sum_{j \in \mathcal{N}} b^k_t(i,j) \cdotp y_t^{jk} }\\
& = \sum_{j \in \mathcal{N}} \frac{ b^k_t(i,j) \cdotp \omega_t^{jk} + b^k_t(i,j) \cdotp \beta_{\omega,t} \cdotp \tilde u_{t+1}^{jk}}{\sum_{l \in \mathcal{N}} b^k_t(i,l) \cdotp y_t^{lk}}\\
& = \sum_{j \in \mathcal{N}} \frac{ (b^k_t(i,j) \cdotp \omega_t^{jk} + b^k_t(i,j) \cdotp \beta_{\omega,t} \cdotp \tilde u_{t+1}^{jk} )/(b^k_t(i,j) \cdotp y_t^{jk})}{1 + \sum_{l \in \mathcal{N}, l\not=j} b^k_t(i,l) \cdotp y_t^{lk} /(b_t(i,j) \cdotp y_t^{jk})}  \\
& = \sum_{j \in \mathcal{N}} \frac{z_t^{jk} + \beta_{\omega,t} \cdotp \tilde u_{t+1}^{jk} / y_t^{jk}}{1 + \sum_{l \in \mathcal{N}, l\not=j} b^k_t(i,l) \cdotp y_t^{lk} /(b^k_t(i,j) \cdotp y_t^{jk})}  \\
& = \sum_{j \in \mathcal{N}} s^k_t(i,j) \cdotp ( z_t^{jk}   + \beta_{\omega,t} \cdotp \epsilon_{t}^{jk} ),
\end{align*}
\normalsize
where $$s^k_t(i,j) = \bigg(1 + \sum_{l \in \mathcal{N}, l\not=j} b^k_t(i,l) \cdotp y_t^{lk} /( b^k_t(i,j) \cdotp y_t^{jk})\bigg)^{-1},$$ 
$\tilde u^i_{t+1} =\tilde \delta_t^i \cdotp \nabla_{z}Q(z_t^i) $, and $ \epsilon_{t}^{ik} =  \tilde u_{t+1}^{ik} / y_t^{ik} $. 
Let $S_t = \sum_{k=1}^K S^k_t \otimes (e_k e_k^\top) $, $z_t = [(z_t^{1})^\top,  \dots, (z_t^{N})^\top ]^\top$, and $\epsilon_t = [(\epsilon_t^{1})^\top,  \dots, (\epsilon_t^{N})^\top ]^\top$, where $S^k_t = [s^k_t(i,j)]_{N \times N}$, $z^i_t = [z_t^{i1},  \dots, z_t^{ik} ]^\top$, and $\epsilon^i_t = [ \epsilon_t^{i1}, \dots, \epsilon_t^{iK}]^\top$. 
Then, 
we have $S_t \cdotp \1 = \1$, and
\begin{equation} \label{equ:pushone_z_s}
z_{t+1} = S_t  \cdotp (z_t + \beta_{\omega,t} \cdotp \epsilon_t).
\end{equation}
From Lemma 1 (b) in \cite{nedic}, we know that 
$\lim_{t \rightarrow \infty} z^i_t = \lim_{t \rightarrow \infty} \langle \widetilde{\omega}_t \rangle$.
Since $ \lim_{t \rightarrow \infty} \beta_{\omega,t} \cdotp \delta_t^i \cdotp \nabla_{z}Q(z_t^i) = \lim_{t \rightarrow \infty} \beta_{\omega,t} \cdotp \delta_t^i \cdotp \phi(s_t,a_t) = 0 $, we have 
$ \lim_{t \rightarrow \infty} \widetilde{\omega}^i_t = \lim_{t \rightarrow \infty} {\omega}^i_t $. 
Thus,
$ \lim_{t \rightarrow \infty} z^i_t = \lim_{t \rightarrow \infty} \langle \widetilde{\omega}_t \rangle =  \lim_{t \rightarrow \infty} \langle {\omega}_t \rangle$, and $\lim_{t \rightarrow \infty} (\langle z_t \rangle - \langle \omega_t \rangle) = 0$.
\hfill $\qed$

Let $y_t = [(y_t^1)^\top, \dots, (y_t^N)^\top]^\top$, where $y_t^i = [y_t^{i1},\dots,y_t^{iK}]^\top$. Then, we have $y_{t+1} = B_t y_t$.
\begin{lemma}\label{lemma:pushone_y_bounded}
Suppose that communication graph $\cG$ is strongly connected. There exists a constant $\alpha > 0$ such that $\alpha \le y_t^{ik} \le N$ for any $i,k,t$ almost surely. 
\end{lemma}

{\em Proof:} 
The existence of a uniform lower bound of $y^{ik}_t$ is a consequence of Lemma~3 in \cite{touri}. 
Since $y^{ik}_0=1$ for all $i,k$ and each $B^k_t$ is column stochastic, $\sum_{i=1}^N y^{ik}_t = \sum_{i=1}^N y^{ik}_0 = N$ for all $t$. It follows that $\alpha \le y_t^{ik} \le N$ for any $i,k,t$.
\hfill $\qed$

\begin{lemma} \label{lemma:pushone_mu_bounded}
Under Assumptions \ref{assum:MDP_Erg} and \ref{assum:Data_Bounded}, the sequence $\{ \mu_t^i \}$ generated as in \eqref{equ:pushone_critic_1} is bounded almost surely.
\end{lemma}

{\em Proof:} 
The proof of the lemma is the same as that of Lemma 5.2 in \cite{kaiqing}.
\hfill $\qed$

\begin{lemma}\label{lemma:pushone_z_bounded} Under Assumptions \ref{assum:MDP_Erg} and \ref{assum:Data_Bounded}-\ref{assum:Linear_Feature_Alg_1}, the sequence $\{z_t^i\}$ is bounded almost surely, i.e., $\sup_t \|{z_t^i}\| < \infty$.
\end{lemma}

{\em Proof:}
Recall that the update of $z$ is 
$ z_{t+1} = S_t  \cdotp (z_t + \beta_{\omega,t} \cdotp \epsilon_t)$ given in \eqref{equ:pushone_z_s}.
Let $h^i(z_t^i,\mu_t^i,y_t^i,s_t,a_t) = \mathbb{E}(\epsilon_t^i|\mathcal{F}_{t,1})$, $M_{t+1}^i = \epsilon_t^i - \mathbb{E}(\epsilon_t^i|\mathcal{F}_{t,1})$. Since the Markov chain $\{(s_t,a_t)\}_{t\ge 0}$ is irreducible and aperidic given policy $\pi_\theta$, we have $\bar h^i(\omega_t^i,\mu_t^i,y_t^i) = \mathbb{E}_{s_t\sim d_\theta,a_t \sim \pi_\theta}[h^i(z_t^i,\mu_t^i,y_t^i,s_t,a_t)] = \Phi^\top D_\theta^{s,a}[\tilde R^i - \1_{N} \otimes \tilde \mu^i_t  + (P^\theta \Phi  - \Phi )\tilde z_t^i]$, where $\tilde R^i = [\frac{R^{i1}}{y_t^{i1}}, \dots, \frac{R^{iK}}{y_t^{iK}},\dots,\frac{R^{i((N-1)K+1)}}{y_t^{i1}}, \dots, \frac{R^{i(NK)}}{y_t^{iK}}]^\top$, $\tilde \mu_t^i = [\frac{\mu_t^{i}}{y_t^{i1}}, \dots, \frac{\mu_t^{i}}{y_t^{iK}}]^\top$ and $\tilde z_t^i = [\frac{z_t^{i1}}{y_t^{i1}}, \dots, \frac{z_t^{iK}}{y_t^{iK}}]^\top$.

From Assumptions \ref{assum:Data_Bounded} and \ref{assum:Linear_Feature_Alg_1},  and Lemmas \ref{lemma:pushone_y_bounded} and \ref{lemma:pushone_mu_bounded}, we know that $\exists K_1, K_2 >0$, s.t. $\|\frac{\phi^k_t}{y_t^{ik}}\|_\infty \le K_1$ and $\|r_{t+1}^i - \mu_t^i\|\le K_2, \forall k,i$. Thus, $\exists K_3 > 0$ such that $ \|\bar h^i(\omega_t^i,\mu_t^i,y_t^i) - h^i(z_t^i,\mu_t^i,y_t^i,s_t,a_t)\|^2 \le K_3 \cdotp (1+\|\omega_t \|^2)$. Moreover, we know $h^i(z_t^i,\mu_t^i,y_t^i,s_t,a_t)$ is Lipschitz continuous in $z_t^i$, and $M_{t+1}^i$ is martingale difference sequence.
Since each $B^k_t$ is column stochastic, it has bounded norm. Thus, by Theorem A.2 in \cite{kaiqing}, it follows that $z_t^i$ is bounded almost surely.
\hfill $\qed$

\begin{lemma} \label{pop:pushone_omega_bounded} Under Assumptions \ref{assum:MDP_Erg} and \ref{assum:Data_Bounded}-\ref{assum:Linear_Feature_Alg_1}, the sequence $\{\omega_t^i\}$ is bounded almost surely, i.e., $\sup_t \|{\omega_t^i}\| < \infty$.
\end{lemma}

{\em Proof:}
From \eqref{equ:pushone_critic_1}, we know that for each entry $k$ in $\omega_t$, $\omega_t^k = z_t^k \cdotp y_t^k$, $k \in \{1,\dots,NK\}$. Moreover, from Lemmas \ref{lemma:pushone_y_bounded} and \ref{lemma:pushone_z_bounded}, $z_t$ and $y_t$ are bounded almost surely. Therefore, it is easy to show that $\omega_t$ is also bounded almost surely.
\hfill $\qed$

We are now in a position to prove Theorem \ref{thm:critic_conv_pushone}.

{\em Proof of Theorem \ref{thm:critic_conv_pushone}:}
Let $\{\mathcal{F}_{t,1}\}$ be the filtration with $\mathcal{F}_{t,1} = \sigma(r_\tau, \mu_\tau, \omega_\tau, z_\tau, y_\tau, s_\tau, a_\tau, B_{\tau-1}, \tau < t)$.  Let $v_t^i = [\mu^i_t, (\omega^i_t)^\top]^\top$ and $v_t = [(v_t^1)^\top, \dots, (v_t^N)^\top]^\top$.
Then, the iteration of $\langle \omega_t \rangle$ has the following form:
\begin{align*}
\langle \omega_{t+1} \rangle & = \frac{1}{N} (\1_N^\top \otimes I_K ) B_{t}( \omega_t  + \beta_{\omega,t} \tilde u_{t+1})\\
& = \frac{1}{N} (\1_N^\top \otimes I_K )( \omega_t  + \beta_{\omega,t} \tilde u_{t+1})\\
& = \langle \omega_t + \beta_{\omega,t} \tilde u_{t+1} \rangle\\
& = \langle \omega_t \rangle + \beta_{\omega,t} \langle \tilde u_{t+1} \rangle\\
& = \langle \omega_t \rangle + \beta_{\omega,t} \langle \tilde \delta_{t} \rangle \cdotp \phi_t.
\end{align*}
Hence, the updates for $ \langle \omega_t \rangle$ and $ \langle \mu_t \rangle$ are
\begin{align}
\langle \mu_{t+1} \rangle &= \langle \mu_t \rangle + \beta_{\omega,t}\cdotp \mathbb{E}(\bar r_{t+1} - \langle \mu_t \rangle | \mathcal{F}_{t,1}) + \beta_{\omega,t} \cdotp  \xi_{t+1,1},\label{equ:pushone_mu}\\ 
\langle \omega_{t+1} \rangle & = \langle \omega_t \rangle + \beta_{\omega,t} \cdotp \mathbb{E}(\langle  \delta_{t} \rangle  \phi_t | \mathcal{F}_{t,1}) + \beta_{\omega,t} \cdotp  \xi_{t+1,2} + \beta_{\omega,t} \cdotp \gamma_{t+1}, \label{equ:pushone_omega}
\end{align}
where $ \xi_{t+1,1} = \bar r_{t+1} - \mathbb{E}(\bar r_{t+1} - \langle \mu_t \rangle | \mathcal{F}_{t,1})$, $\xi_{t+1,2} = \langle  \delta_{t} \rangle  \phi_t - \mathbb{E}(\langle  \delta_{t} \rangle  \phi_t | \mathcal{F}_{t,1})$, and $\gamma_{t+1} = \langle \tilde \delta_{t} \rangle \phi_t - \langle  \delta_{t} \rangle \phi_t $.

Note that $ \mathbb{E}(\bar r_{t+1} - \langle \mu_t \rangle | \mathcal{F}_{t,1})$ is Lipschitz continuous in $ \langle \mu_t \rangle$, and that $\mathbb{E}(\langle  \delta_{t} \rangle  \phi_t | \mathcal{F}_{t,1})$ is Lipschitz continuous in both $ \langle \omega_t \rangle$ and $ \langle \mu_t \rangle$. Moreover, $ \xi_{t+1,1}$ and $ \xi_{t+1,2}$ are martingale differences sequences. 
From Lemmas \ref{lemma:pushone_y_bounded} and \ref{pop:pushone_omega_bounded}, $\{\gamma_{t}\}$ is a bounded random sequence with $\gamma_{t} \rightarrow 0$ as $t \rightarrow \infty$ almost surely.

From Theorem B.2 in \cite{kaiqing}, the following ODE captures the asymptotic behavior of \eqref{equ:pushone_mu} and \eqref{equ:pushone_omega}:
\begin{align}
\nonumber \langle \dot{v} \rangle &= 
\left[
\begin{array}{lr}  
\langle \dot{\mu} \rangle\\
\langle \dot{\omega} \rangle
\end{array}  
\right] \\
& =
\left[
\begin{array}{cc}  
-1 & 0\\
-\Phi^\top D_\theta^{s,a}\1_{NK} & \Phi^\top D_\theta^{s,a}(P^\theta -I_{NK})\Phi
\end{array}  
\right]
\left[
\begin{array}{lr}  
\langle {\mu} \rangle\\
\langle {\omega} \rangle
\end{array}  
\right] +
\left[
\begin{array}{lr}  
J(\theta)\\
\Phi^\top D_\theta^{s,a}\bar{R}
\end{array}  
\right] \label{equ:pushone_ODE}
\end{align}
where $D_\theta^{s,a} = \diag[d_\theta(s)\cdotp \pi_\theta(s,a),s\in \mathcal{S},a\in\mathcal{A}].$
From the proof of Theorem 4.6 in \cite{kaiqing}, the ODE \eqref{equ:pushone_ODE} is globally asymptotically stable and has its equilibrium satisfying
\begin{equation} \label{euq:pushone_solution_omega_mu}
\left\{
\begin{array}{l}
\langle {\mu} \rangle  = J(\theta)\\
\Phi^\top D_\theta^{s,a}[\bar{R} - \langle {\mu} \rangle \1_{NK} + P^\theta \Phi \langle {\omega} \rangle - \Phi \langle {\omega} \rangle ] = 0. 
\end{array}
\right.
\end{equation}
Note that the solution for $\langle \mu \rangle$ at equilibrium is $J(\theta)$, and that the solution for $\langle \omega \rangle$ has the form $\omega_\theta + l v$ with any $l\in\mathbb{R}$ and $v\in\mathbb{R}^K$ such that $\phi v=\1_{K}$, where $\omega_\theta$ follows that $\Phi^{\top}\Db^{s,a}_{{\theta}}\big[T^Q_{\theta}(\Phi \omega_\theta)-\Phi \omega_\theta\big]={0}$. Moreover, $\phi v \not= \1_{K}$ by Assumption \ref{assum:Linear_Feature_Alg_1}, so $\omega_\theta$ is the unique solution, which implies that $\lim_t \langle \omega_t \rangle = \omega_\theta$. Combining the above facts with Lemma~\ref{lemma:pushone_converge}, we conclude that $ \lim_{t}z_t^i = \omega_\theta $.

As for the actor step convergence, the proof is the same as that of Theorem 4.7 in \cite{kaiqing}. 
\hfill $\qed$

\section{Conclusions} \label{ending}

This paper has proposed a communication-efficient distributed reinforcement learning algorithm.
We have shown that the algorithm allows each agent to only transmit two scalar-valued variables, one of which is an independently selected entry of the agent's estimate vector, at each time, and works for any strongly connected graph, which significantly reduces communication cost at one time compared with the algorithms in \cite{kaiqing}. It is fairly  straightforward to extend the algorithm and its convergence result to the case where each agent transmits more than one entry of its estimate vector. It is also fairly straightforward to extend the proposed algorithm to an asynchronous case without communication delays as was done in \cite{acc12}. In the case when communication delays are taken into account, a modified version of the algorithm here is expected to solve the problem using the idea in \cite{pushdelay}. Future directions of this work include comparison of total communication cost with the algorithms in \cite{kaiqing}, extension to time-varying communication graphs, and development of asynchronous versions of the algorithm. 

\section{Appendix}

{\em Proof of Proposition~\ref{yixuan}:}
(1) Since all $C^k_t(t)$ satisfy condition 1) in Assumption \ref{assum:MARL_weighted_matrix}, it follows that 
\begin{align*}
    \bar C_t \cdotp \1_{NK} 
    & = \sum_{k=1}^K  (C^k_{t} \otimes (e_k e_k^\top)) \cdotp (\1_{N} \otimes \1_{K}) \\
    & = \sum_{k=1}^K  (C^k_{t} \cdotp \1_{N})  \otimes (e_k e_k^\top \cdotp \1_{K}) = \1_{NK},\\
 \1^\top_{NK} \mathbb{E}[\bar C_t] 
 & = \sum_{k=1}^K (\1^\top_{N}\otimes \1^\top_{K}) (\mathbb{E}[C^k_{t}] \otimes (e_k e_k^\top)) \\
 & = \sum_{k=1}^K (\1^\top_{N}\cdotp \mathbb{E}[C^k_{t}])\otimes(\1^\top_{K}\cdotp (e_k e_k^\top)) = \1^\top_{NK}.
\end{align*}
From the definition of $\bar C_t$, for any entry $(i,j)$ of $C^k_t$, $\bar c_t((i-1)K+k,(j-1)K+k) = c^k_t(i,j)$.
Since for any $c^k_t(i,j) > 0$, $c_t^k(i,j) \ge \eta, \forall k \in \{1,\dots,K\},i,j\in\{1,\dots,N\}$,  each entry of $\bar C_t$ also satisfies condition 1).

(2) From the definition of $\bar C_t$, since all $C^k_t$ satisfy condition 2) in Assumption \ref{assum:MARL_weighted_matrix}, so does $\bar C_t$.

(3) 
First note that
\begin{align*}
& \mathbb{E}[ \bar C_t^\top \cdotp (I_{N}-\1_{N}\1_{N}^\top/N)  \otimes I_{K} \cdotp \bar C_t  ]\\
& = \mathbb{E}[(\sum_{k=1}^K  C^k_{t} \otimes (e_k e_k^\top))^\top \cdotp (I_{N}-\1_{N}\1_{N}^\top/N) \otimes (\sum_{k=1}^K e_k e_k^\top) \cdotp (\sum_{k=1}^K C^k_{t} \otimes (e_k e_k^\top))  ]\\
& = \mathbb{E}[(\sum_{k=1}^K (C^k_{t})^\top \otimes (e_k e_k^\top)) \cdotp (\sum_{k=1}^K (I_{N}-\1_{N}\1_{N}^\top/N)  \otimes (e_k e_k^\top)) \cdotp (\sum_{k=1}^K C^k_{t} \otimes (e_k e_k^\top))  ]\\
& = \mathbb{E}[(\sum_{k=1}^K  ((C^k_{t})^\top \cdotp (I_{N}-\1_{N}\1_{N}^\top/N) \cdotp C^k_{t}) \otimes (e_k e_k^\top))].
\end{align*}
There exists a permutation matrix $D$ such that 
\begin{align*}
    & D^\top\cdotp \mathbb{E}[(\sum_{k=1}^K ((C^k_{t})^\top \cdotp (I_{N}-\1_{N}\1_{N}^\top/N) \cdotp C^k_{t}) \otimes (e_k e_k^\top))] \cdotp D \\
    & = \mathbb{E}[\diag\{ ((C^1_{t})^\top (I_{N}-\1_{N}\1_{N}^\top/N) C^1_{t}), \dots,((C^K_{t})^\top (I_{N}-\1_{N}\1_{N}^\top/N)  C^K_{t}) \}].
\end{align*} 
Thus, the spectral norm of $\mathbb{E}[(\sum_{k=1}^K ((C^k_{t})^\top \cdotp (I_{N}-\1_{N}\1_{N}^\top/N) \cdotp C^k_{t}) \otimes (e_k e_k^\top))]$ is the same as that of $\mathbb{E}[\diag\{ ((C^1_{t})^\top (I_{N}-\1_{N}\1_{N}^\top/N) C^1_{t}), \dots,((C^K_{t})^\top (I_{N}-\1_{N}\1_{N}^\top/N)  C^K_{t}) \}]$.
Since the spectral norm of \\
$\mathbb{E}[(C^k_{t})^\top (I_{N}-\1_{N}\1_{N}^\top/N) C^k_{t})]$ is strictly less than one for all $k$, so is the spectral norm of 
$$\mathbb{E}[\diag\{ ((C^1_{t})^\top (I_{N}-\1_{N}\1_{N}^\top/N) C^1_{t}), \dots, ((C^K_{t})^\top (I_{N}-\1_{N}\1_{N}^\top/N)  C^K_{t}) \}].$$ 
Thus, the spectral norm of $\mathbb{E}[\bar C_t^\top \cdotp (I_{N}-\1_{N}\1_{N}^\top/N)  \otimes I_{K} \cdotp \bar C_t  ]$ is strictly less than one. 

(4) From the definition of $\bar C_t$, it is easy to see that $\bar C_t$ is conditionally independent of $r_{t+1}^i$ for any $i$ as all $C^k_t(t)$ are. 
This completes the proof.
\hfill $\qed$

\bibliographystyle{unsrt}
\bibliography{rl,AC_over_networks}

\end{document}